\documentclass[11pt]{article}
\usepackage{nodalida2021}
\usepackage{times}
\usepackage{hyperref}
\usepackage{latexsym}
\usepackage{graphicx}
\usepackage{booktabs}
\usepackage{multirow}
\usepackage{xspace}
\usepackage{amsmath}
\usepackage{caption}
\usepackage{subcaption}

\newcommand\NorecFine{{NoReC$_\text{\textit{fine}}$}\xspace}
\newcommand\NorecNeg{{NoReC$_\text{\textit{neg}}$}\xspace}
\newcommand\NorecSentence{{NoReC$_\text{\textit{sentence}}$}\xspace}

\newcommand{\F}{F$_1$\xspace}

\aclfinalcopy % Uncomment this line for the final submission

\title{Large-Scale Contextualised Language Modelling for Norwegian}

\author{Andrey Kutuzov, Jeremy Barnes, Erik Velldal,\\[0.5ex] \bf Lilja Øvrelid and Stephan Oepen\\[1ex]
  University of Oslo\\
  Department of Informatics\\
  Language Technology Group\\[1ex]
  $\{\,$\texttt{andreku}$\,|\,$\texttt{jeremycb}$\,|\,$\texttt{erikve}$\,|\,$\texttt{liljao}$\,|\,$\texttt{oe}$\,\}\,$\texttt{@ifi.uio.no}}

\date{}

\begin{document}
\maketitle
\begin{abstract}
We present the ongoing NorLM initiative to support the creation and use of very large contextualised language models for Norwegian (and in principle other Nordic languages),
including a ready-to-use software environment, as well as an experience
report for data preparation and training. 
This paper introduces the first  large-scale monolingual language models for Norwegian, based on both the ELMo and BERT frameworks. In addition to detailing the training process, we present contrastive benchmark results on a suite of NLP tasks for Norwegian. 

For additional background and access to the data, models, and software, please see:
\begin{center}
  \vspace*{-2ex}
  \url{http://norlm.nlpl.eu}
\end{center}
\end{abstract}

\section{Introduction} \label{sec:intro}

In this work, we present \textit{NorLM}, an ongoing community initiative and emerging collection of large-scale contextualised language models for Norwegian.  
We here introduce the NorELMo and NorBERT models, that have been trained on around two billion tokens of running Norwegian text. We describe the training procedure and compare these models with the multilingual mBERT model \cite{devlin-etal-2019-bert}, as well as an additional Norwegian BERT model developed contemporaneously, with some interesting differences in training data and setup. 
We report results over a number of Norwegian benchmark datasets, addressing a broad range of diverse NLP tasks: part-of-speech tagging, negation resolution, sentence-level and fine-grained sentiment analysis and named entity recognition (NER). 

 All the models are publicly available for download 
 from the Nordic Language Processing Laboratory (NLPL) Vectors Repository\footnote{\url{http://vectors.nlpl.eu/repository}} with a CC BY 4.0 license. They are also accessible locally, together with the training and supporting software, on the two national superclusters Puhti and Saga, in Finland and Norway, respectively, which are available to university NLP research groups in Northern Europe through the Nordic Language Processing Laboratory (NLPL).\footnote{\url{http://www.nlpl.eu}} The NorBERT model is in addition served via the Huggingface Transformers model hub.\footnote{\url{https://huggingface.co/ltgoslo/norbert}}

NorLM is a joint effort of the projects EOSC-Nordic (European Open Science Cloud) and SANT (Sentiment Analysis for Norwegian), coordinated by the Language Technology Group (LTG) at the University of Oslo.
The goal of this work is to provide these models and supporting tools for researchers and developers in Natural Language Processing (NLP) for the Norwegian language. We do so in the hope of facilitating scientific experimentation with and practical applications of state-of-the-art NLP architectures, as well as to enable others to develop their own large-scale models, for example for domain- or application-specific tasks, language variants, or even other languages than Norwegian. 
Under the auspices of the NLPL use case in EOSC-Nordic, we are also coordinating with colleagues in Denmark, Finland, and Sweden on a collection of large contextualised language models for the Nordic languages, including language variants or related groups of languages, as linguistically or technologically appropriate. 

\section{Background}
\label{sec:background}

\paragraph{Bokmål and Nynorsk} 
There are two official standards for written Norwegian; \emph{Bokmål}, the main variety, and \emph{Nynorsk}, used by 10--15\% of the Norwegian population. Norwegian language legislation specifies that minimally 25\% of the written public service information should be in Nynorsk. While the two varieties are closely related, there can also be relatively large differences lexically (though often with a large degree of overlap on the character-level still). Several previous studies have indicated that joint modeling of Bokmål and Nynorsk works well for many NLP tasks, like tagging and parsing \cite{VelOvrHoh17} and NER \cite{JorAasHus20}. The contextualised language models presented in this paper are therefore trained jointly on both varieties, but with the minority variant Nynorsk represented by comparatively less data than Bokmål (reflecting the natural usage). 

\paragraph{Datasets} 
For all our models presented below, we used the following training corpora:

 \begin{enumerate}
  \item Norsk Aviskorpus (NAK), a collection of Norwegian news texts\footnote{\url{https://www.nb.no/sprakbanken/ressurskatalog/oai-nb-no-sbr-4/}} (both Bokmål and Nynorsk) from 1998 to 2019; 1.7 billion words;
  \item Bokmål Wikipedia dump from September 2020; 160 million words;
  \item Nynorsk Wikipedia dump from September 2020; 40 million words.
 \end{enumerate}

The corpora contain ordered sentences (which is important for BERT-like models, because one of their training tasks is next sentence prediction). In total, our training corpus comprises about two billion (1,907,072,909) word tokens in 203 million (202,802,665) sentences.

We conducted the following pre-processing steps:
\begin{enumerate}
    \item Wikipedia texts were extracted from the dumps using the \texttt{segment\_wiki} script from the Gensim project \cite{rehurek_lrec}.
    \item For the news texts from Norwegian Aviskorpus, we performed de-tokenization and conversion to UTF-8 encoding, where required.
    \item The resulting corpus was sentence-segmented using Stanza \cite{qi-etal-2020-stanza}. We left blank lines between documents (and sections in the case of Wikipedia) so that the `next sentence prediction' task of BERT does not span between documents. 
\end{enumerate}

\section{Prerequisites: software and computing}

Developing very large contextualised language models is no small challenge, both in terms of engineering sophistication and computing demands.
Training ELMo- and in particular BERT-like models presupposes access to specialised hardware -- graphical processing units (GPUs) -- over extended periods of time.
Compared to the original work at Google or to our sister initiative at the National Library of Norway (see below), our two billion tokens in Norwegian training data can be characterised as moderate in size.

Nevertheless, training a single NorBERT model requires close to one full year of GPU utilisation, which through parallelization over multiple compute nodes, each featuring four GPUs, could be completed in about three weeks of wall clock time.
At this scale, premium software efficiency and effective parallelization are prerequisites, not only to allow repeated incremental training and evaluation cycles to complete in practical intervals, but equally so for cost-efficient utilisation of scarce, shared computing resources and, ultimately, a shred of environmental sustainability.

To prepare the NorLM software environment, we have teamed up with support staff at the Norwegian national e-infrastructure provider, Uninett Sigma2, and developed a fully automated and modularised installation procedure using the EasyBuild framework (\url{https://easybuild.io}).
All necessary tools are compiled from source with the right set of hardware-specific optimizations and platform-specific optimised libraries for basic linear algebra (`math kernels') and communication across multiple compute nodes.

This approach to software provisioning makes it possible to (largely) automatically create fully parallel training and experimentation environments on multiple computing infrastructures -- in our work to date two national HPC superclusters, in Norway and Finland, but in principle just as much any suitable local GPU cluster.
In our view, making available both a ready-to-run software environment on Nordic national e-infrastructures, where university research groups typically can gain no-cost access, coupled with the recipe for recreating the environment on other HPC systems, may contribute to `democratising' large-scale NLP research; if nothing else, it eliminates dependency on commercial cloud computing services.

\section{Related work} \label{sec:related}

Large-scale deep learning language models (LM) are important components of current NLP systems. They are often based on BERT (Bidirectional Encoder Representations from Transformers) \cite{devlin-etal-2019-bert} and other contextualised architectures.
A number of language-specific initiatives have in recent years released monolingual versions of these models for a number of languages \cite{fares-etal-2017-word,kutuzov-kuzmenko-2017-building,virtanen2019multilingual,deV:Wie:Cra:2019,Ulc:Rob:Sik:2020,Kou:Cha:Mal:2020,Ngu:Quo:Ngu:2020,Far:Gha:Far:2020,malmsten2020playing}. For our purposes, the most important such previous training effort is that of \newcite{virtanen2019multilingual} on creating a BERT model for Finnish --  FinBERT\footnote{\url{https://github.com/TurkuNLP/FinBERT}} -- as our training setup for creating NorBERT builds heavily on this; see Section~\ref{sec:norbert} for more details. 

Many low-resource languages do not have dedicated monolingual large-scale language models, and instead resort to using a multilingual model, such as Google's multilingual BERT model -- mBERT -- which was trained on data that also included Norwegian. Up until the release of the models described in the current paper, mBERT was the only BERT-instance that could be used for Norwegian.\footnote{A BERT model trained on Norwegian data was published at \url{https://github.com/botxo/nordic_bert} in the beginning of 2020. However, the vocabulary of this model seems to be broken, and to the best of our knowledge nobody has achieved any meaningful results with it.}

Another widely used architecture for contextualised LMs is Embeddings From Language Models or ELMo \cite{peters-etal-2018-deep}. The \textit{ElmoForManyLangs} initiative \cite{che-etal-2018-towards} trained and released monolingual ELMo models for a wide range of different languages, including Norwegian (with separate models for Bokmål and Nynorsk). However, these models were trained on very modestly sized corpora of 20 million words for each language (randomly sampled from Wikipedia dumps and Common Crawl data). 

In a parallel effort to that of the current paper, the AI Lab of the National Library of Norway, through their Norwegian Transformer Model (NoTraM) project, has released a Norwegian BERT (Base, cased) model dubbed NB-BERT \cite{KumJavWet21}.\footnote{\url{https://github.com/NBAiLab/notram}} The model is trained on the Colossal Norwegian Corpus, reported to comprise close to 18,5 billion words (109.1 GB of text). 

In raw numbers, this is about ten times more than the corpus we use for training the NorLM models. However, the vast majority of this is from OCR'ed historical sources, which is bound to introduce at least some noise. In Section~\ref{sec:tasks} below, we demonstrate that in some NLP tasks, a language model trained on less (but arguably cleaner) data can outperform a model trained on larger but noisy corpora.

\section{NorELMo} \label{sec:norelmo}

NorELMo is a set of bidirectional recurrent ELMo language models trained from scratch on the Norwegian corpus described in Section~\ref{sec:intro}. They can be used as a source of contextualised token representations for various Norwegian natural language processing tasks. As we show below, in many cases, they present a viable alternative to Transformer-based models like BERT. Their performance is often only marginally lower, while the compute time required to adapt the model to the task at hand can be an order of magnitude less on identical hardware.   

Currently we present two models, with more following in the future:
\begin{enumerate}
 \item \textbf{NorELMo$_{30}$}: 30,000 most frequent words in the vocabulary
 \item \textbf{NorELMo$_{100}$}: 100,000 most frequent words in the vocabulary
\end{enumerate}

Note that independent of the vocabulary size, both NorELMo$_{30}$ and NorELMo$_{100}$ can process arbitrary word tokens, due to the ELMo architecture (where the first CNN layer converts input strings to non-contextual word embeddings). Thus, the size of the vocabulary controls only the number of words used as targets for the language modelling task in the course of training. Supposedly, the model with a larger vocabulary is more effective in treating less frequent words at the cost of being less effective with more frequent words.

Each model was trained for 3 epochs with batch size 192. We employed a version of the original ELMo training code from \newcite{peters-etal-2018-deep} updated to work better with the recent TensorFlow versions.
All the hyperparameters were left at their default values, except the LSTM dimensionality reduced to 2,048 from the default 4,096 (in our experience, this rarely influences performance). Training of each model took about 100 hours on four NVIDIA P100 GPUs.

These are the first ELMo models for Norwegian trained on a large corpus. As has already been mentioned, the Norwegian ELMo models from the \textit{ElmoForManyLangs} project \cite{che-etal-2018-towards} were trained on very small corpora samples and seriously under-perform on semantic-related NLP  tasks, although they can yield impressive results on POS tagging and syntactic parsing \cite{zeman-etal-2018-conll}. In addition, they were trained with custom code modifications and can be used only with the custom \textit{ElmoForManyLangs} library. On the other hand, our NorELMo models are fully compatible both with the original ELMo implementation by \newcite{peters-etal-2018-deep} and with the more modern \textit{simple\_elmo} Python library provided by us.\footnote{\url{https://pypi.org/project/simple-elmo/}} 

The vocabularies are published together with the models. For different tasks, different models can be better, as we show below. The published packages contain both TensorFlow checkpoints (for possible fine-tuning, if need be) and model files in the standard Hierarchical Data Format (HDF5) for easier inference usage. In addition, we have setup ELMoViz, a demo web service to explore Norwegian ELMo models.\footnote{\url{http://vectors.nlpl.eu/explore/embeddings/en/contextual/}}

\section{NorBERT} \label{sec:norbert}

\begin{figure*}
    \centering
    \includegraphics[width=0.45\linewidth]{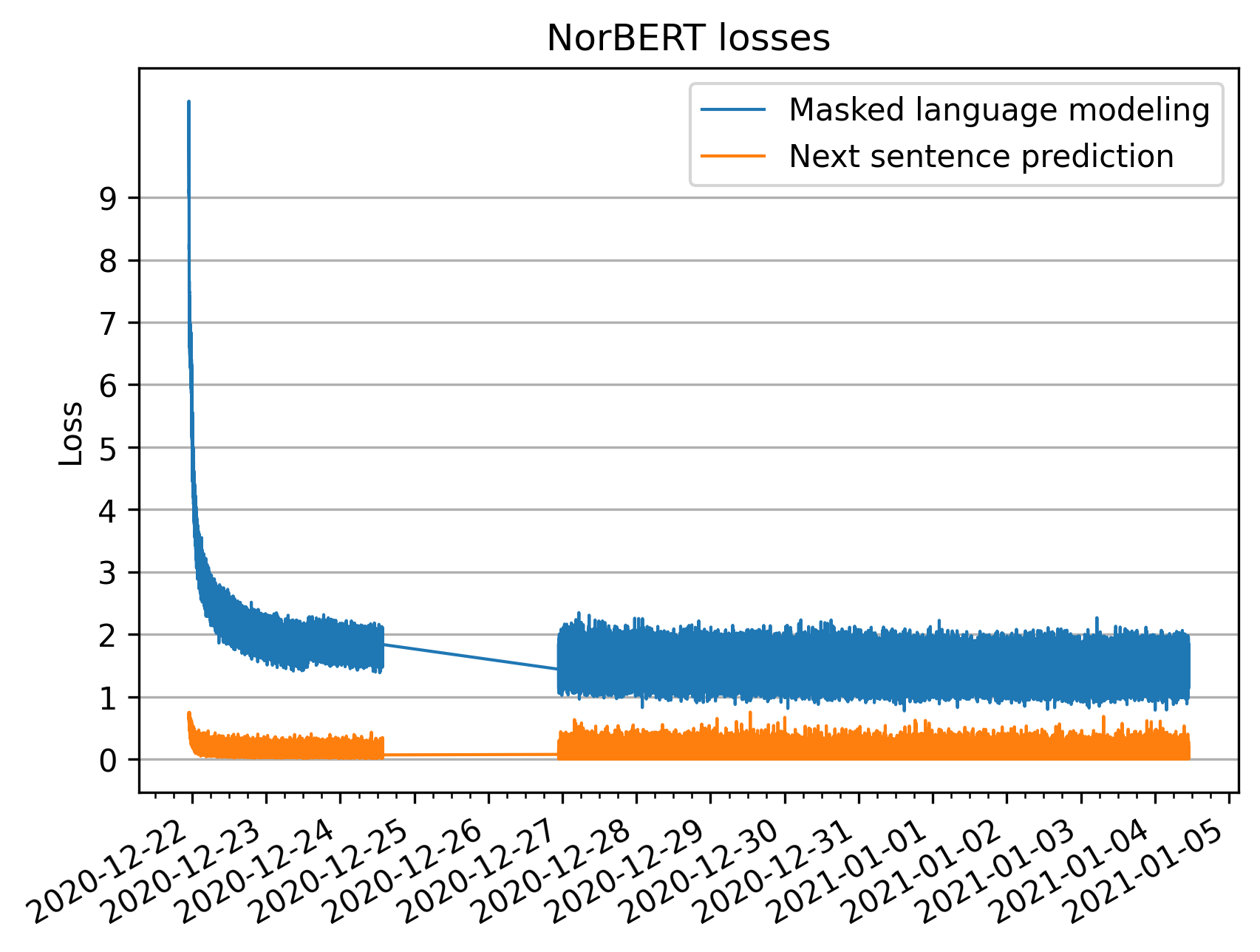}
    \includegraphics[width=0.45\linewidth]{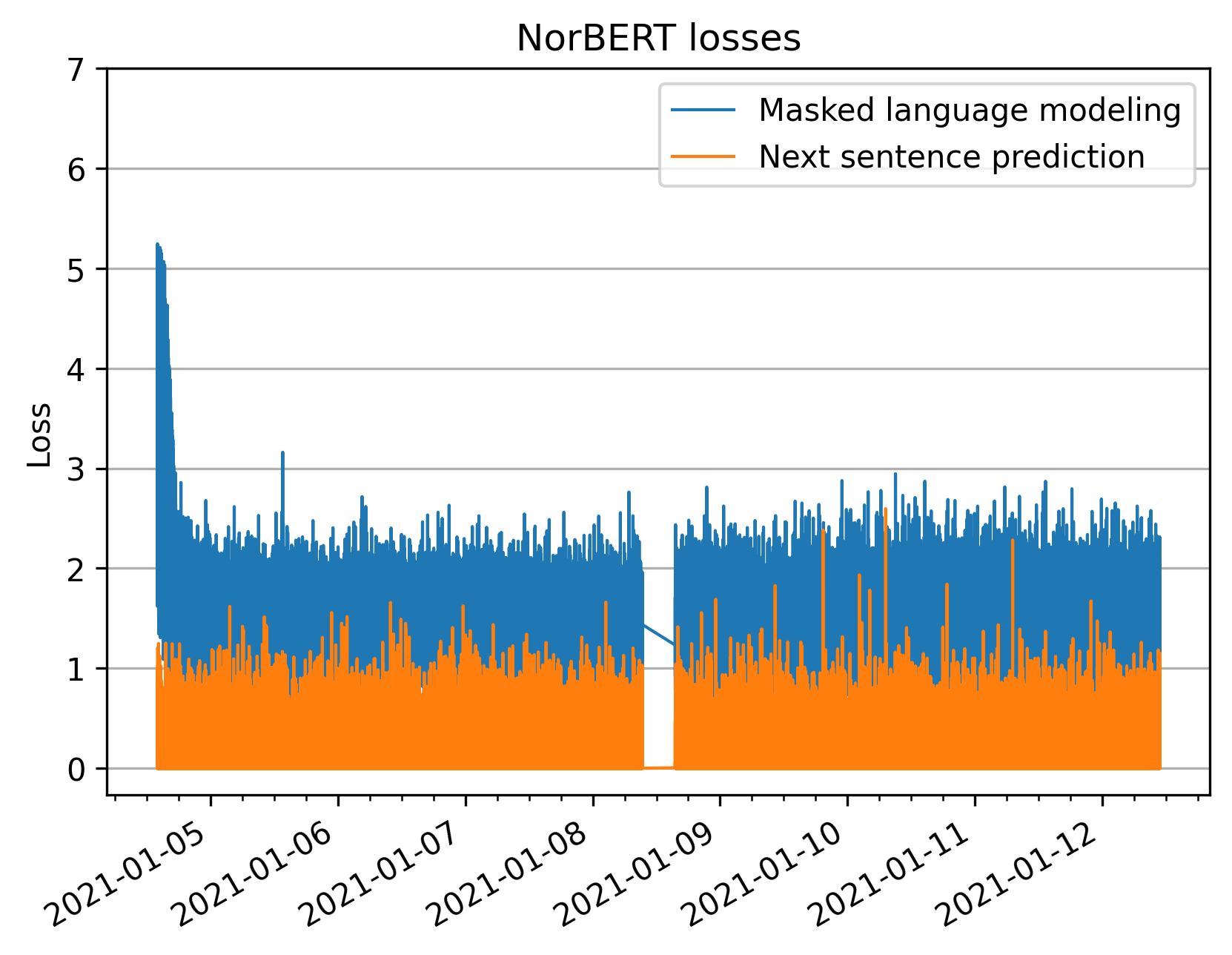}
    \caption{NorBERT loss plots at the Phase 1 (left) and Phase 2 (right).}
    \label{fig:bert_loss}
\end{figure*}

Our NorBERT model is trained from scratch for Norwegian, and can be used in exactly the same way as any other BERT-like model. The NorBERT training setup heavily builds on prior work on FinBERT conducted at the University of Turku \cite{virtanen2019multilingual}. 

NorBERT features a custom WordPiece vocabulary which is case-sensitive and includes accented characters. It has much better coverage of Norwegian words than the mBERT model or NB-BERT (which uses the same vocabulary as mBERT). This is clearly seen on the example of the tokenization performed by both for the Norwegian sentence `\textit{Denne gjengen håper at de sammen skal bidra til å gi kvinnefotballen i Kristiansand et lenge etterlengtet løft}'

 \begin{itemize}
  \item \textbf{mBERT/NB-BERT:} `Denne g \#\#jeng \#\#en h \#\#å \#\#per at de sammen skal bid \#\#ra til å gi k \#\#vinne \#\#fo \#\#t \#\#ball \#\#en i Kristiansand et lenge etter \#\#len \#\#gte \#\#t l \#\#ø \#\#ft'
  \item \textbf{NorBERT:} `Denne gjengen håper at de sammen skal bidra til å gi kvinne \#\#fotball \#\#en i Kristiansand et lenge etterl \#\#engt \#\#et løft'
 \end{itemize}

NorBERT tokenization splits the sentence into pieces which much better reflect the real Norwegian words and morphemes (cf. `\textit{k vinne fo t ball en}' versus `\textit{kvinne fotball en}'). We believe this to be extremely important for more linguistically-oriented studies, where it is critical to deal with words, not with arbitrarily fragmented pieces (even if they are well-performing in practical tasks).

The vocabulary for the model is of size 30,000. It is  much less than the 120,000 of mBERT, but it is compensated by these entities being almost exclusively Norwegian. The vocabulary was generated from raw text, without, e.g., separating punctuation from word tokens. This means one can feed raw text into NorBERT. 

For the vocabulary generation, we used the SentencePiece algorithm \cite{kudo-2018-subword} and Tokenizers library.\footnote{\url{https://github.com/huggingface/tokenizers}} The resulting Tokenizers model was converted to the standard BERT WordPiece format.  The final vocabulary contains several thousand unused wordpiece slots which can be filled in with task-specific lexical entries for further fine-tuning by future NorBERT users.

\subsection{Training technicalities}

NorBERT corresponds in its configuration to the Google's Bert-Base Cased for English, with 12 layers and hidden size 768 \cite{devlin-etal-2019-bert}. We used the standard masked language modeling and next sentence prediction losses with the LAMB optimizer \cite{you2020large}.
The model was trained on the Norwegian academic HPC system called Saga. Most of the time the training process was distributed across 4 compute nodes and 16 NVIDIA P100 GPUs. Overall, it took approximately 3 weeks (more than 500 hours). 

Similar to \newcite{virtanen2019multilingual}, we employed the BERT implementation by NVIDIA\footnote{\url{https://github.com/NVIDIA/DeepLearningExamples/tree/master/TensorFlow/LanguageModeling/BERT}, version 20.06.08}, which allows fast multi-node and multi-GPU training.

We made minor changes to this code, mostly to adapt it to the newer TensorFlow versions. All these patches and the utilities we used at the pre-processing, training and evaluation stages are published in our GitHub repository.\footnote{\url{https://github.com/ltgoslo/NorBERT}} 
Instructions to reproduce the training setup with the EasyBuild software build and installation framework are also available.\footnote{\url{http://wiki.nlpl.eu/index.php/Eosc/pretraining/nvidia}}

\subsection{Training workflow}

Phase 1 (training with maximum sequence length of 128) was done with batch size 48 and global batch size 48*16=768. Since one global batch contains 768 sentences, approximately 265,000 training steps constitute 1 epoch (one pass over the whole corpus). We have done 3 epochs: 795,000 training steps.

Phase 2 (training with maximum sequence length of 512) was done with batch size 8 and global batch size 8*16=128. We aimed at mimicking the original BERT in that at Phase 2 the model should see about 1/9 of the number of sentences seen during Phase 1. Thus, we needed about 68 million sentences, which at the global batch size of 128 boils down to 531,000 training steps more.

The loss plots are shown in Figure~\ref{fig:bert_loss} (the training was on pause on December 25 and 26, since we were solving problems with mixed precision training). Full logs are available at the GitHub repository.

\section{Evaluation}
\label{sec:tasks}

This section presents benchmark results across a range of different tasks. We compare NorELMO and NorBERT to both mBERT and to the recently released  NB-BERT model described in Section~\ref{sec:related}. Where applicable, we show separate evaluation results for Bokmål and Nynorsk. 
Below we first provide an overview of the different tasks and the corresponding classifiers that we train, before turning to discuss the results. 

\subsection{Task descriptions}

We start by briefly describing each task and associated dataset, in addition to the architectures we use.  The sentence counts for the different datasets and their train, dev. and test splits are provided in Table~\ref{tab:testsets}.

\begin{table}
    \centering
    \begin{tabular}{@{}lrrr@{}}
    \toprule
    Task &  Train &  Dev & Test \\
    \midrule
    POS Bokmål & 15,696 & 2,409 & 1,939 \\
    POS Nynorsk & 14,174 & 1,890 & 1,511 \\
    NER Bokmål & 15,696 & 2,409 & 1,939  \\
    NER Nynorsk & 14,174 & 1,890 & 1,511 \\
    Sentence-level SA  & 2,675 & 516  & 417 \\
    Fine-grained SA & 8,543 & 1,531 & 1,272 \\
    Negation & 8,543 & 1,531 & 1,272 \\
    \bottomrule
    \end{tabular}
    \caption{Number of sentences in the training, development, and test splits in the datasets used for the evaluation tasks.}
    \label{tab:testsets}
\end{table}

\paragraph{Part-of-speech tagging} The Norwegian Dependency Treebank (NDT) \cite{Sol:Skj:Ovr:14} includes annotation of POS tags for both Bokmål and Nynorsk. NDT has also been converted to the Universal Dependencies format \cite{OvrHoh16, VelOvrHoh17} and this is the version we are using here (for UD 2.7) for predicting UPOS tags. 

We use a typical sequence labelling approach with the BERT models, adding a linear layer after the final token representations and  taking the softmax to get token predictions. We fine-tune all parameters for 20 epochs, using a learning rate of 2e-5, a training batch size of 8, max length of 256, and keep the best model on the development set. ELMo models were not fine-tuned, following the recommendations from \newcite{peters-etal-2019-tune}. Instead we trained a simple neural classifier (a feed forward network with one hidden layer of size 128, ReLU non-linear activation function and dropout), using ELMo token embeddings as features. The random seed has been kept fixed all the time. Models are evaluated on accuracy.

\paragraph{Named entity recognition} %(Andrey, Lilja) 
The NorNE\footnote{\url{https://github.com/ltgoslo/norne}} dataset annotates the UD-version of NDT with a rich set of entity types \cite{JorAasHus20}.  
The evaluation metrics here is `strict' micro \F,  requiring both the correct entity type and exact match of boundary surface string. We predict  8 entity types: Person (PER), Organisation (ORG), Location (LOC), Geo-political entity, with a locative sense (GPE-LOC), Geo-political entity, with an organisation sense (GPE-ORG), Product (PROD), Event (EVT), Nominals derived from names (DRV). The evaluation is done using the code for the SemEval'13 Task 9\footnote{\url{https://github.com/davidsbatista/NER-Evaluation}}.

We cast the named entity recognition problem as a sequence labelling task, using a BIO label encoding. For the BERT-based models, we solve it by fine-tuning the pre-trained model on the NorNE dataset for 20 epochs with early stopping and batch size 32. The resulting model is applied to the test set.

For ELMo models, we infer contextualised token embeddings (averaged representations across all 3 layers) for all words. Then, these token embeddings are fed to a neural classifier with dropout, identical to the one we used for POS tagging earlier. This classifier is also trained for 20 epochs with early stopping and batch size 32.

\paragraph{Fine-grained sentiment analysis} \NorecFine is a dataset\footnote{\url{https://github.com/ltgoslo/norec_fine}} comprising a subset of the Norwegian Review Corpus  \citep[NoReC;][]{VelOvrBer18} annotated for sentiment holders, targets, expressions, and polarity, as well as the relationships between them \cite{ovrelid-etal-2020-fine}. 
We here cast the problem as a graph prediction task and train a graph parser \cite{dozat-manning-2018-simpler,kurtz-etal-2020-end} to predict sentiment graphs. The parser creates token-level representations which is the concatenation of a word embedding, POS tag embedding, lemma embedding, and character embedding created by a character-based LSTM. We further augment these representations with contextualised embeddings from each model. Models are trained for 100 epochs, keeping the best model on development \F. For span extraction (holders, targets, expressions), we evaluate token-level \F, and the common Targeted \F metric, which requires correctly extracting a target (strict) and its polarity. We also evaluate Labelled and Unlabelled \F, which correspond to Labelled and Unlabelled Attachment in dependency parsing. Finally, we evaluate on Sentiment Graph \F (\textbf{S\F}) and Non-polar Sentiment Graph \F (\textbf{NS\F}. S\F requires predicting all elements (holder, target, expression, polarity) and their relationships (NS\F removes the polarity). A true positive is defined as an exact match at graph-level, weighting the overlap in predicted and gold spans for each element, averaged across all three spans. For precision we weight the number of correctly predicted tokens divided by the total number of predicted tokens (for recall, we divide instead by the number of gold tokens). We allow for empty holders and targets.

\paragraph{Sentence-level binary sentiment classification} We further evaluate on the task of sentence-level binary (positive or negative) polarity classification, using labels that we derive from \NorecFine described above. We create the dataset for this by aggregating the fine-grained annotations to the sentence-level,  removing sentences with mixed or no sentiment. The resulting dataset, \NorecSentence, is made publicly available.\footnote{\url{https://github.com/ltgoslo/norec_sentence}} For the BERT models, we use the \texttt{[CLS]} embedding of the last layer as a representation for the sentence and pass this to a softmax layer for classification. We fine-tune the models in the same way as for the POS tagging task, training the models for 20 epochs and keeping the model that performs best on the development data. For ELMo models, we used a BiLSTM with global max pooling, taking ELMo token embeddings from the top layer as an input. The evaluation metric is macro \F.

\paragraph{Negation detection}
Finally, the \NorecFine dataset has recently been annotated with negation cues and their corresponding in-sentence scopes  \cite{MaeBarKur21}. The resulting dataset is dubbed \NorecNeg.\footnote{\url{https://github.com/ltgoslo/norec_neg}} We use the same graph-based modeling approach as described for fine-grained sentiment above. We evaluate on the same metrics as in the *SEM 2012 shared task \citep{Morante2012}: cue-level \F (CUE), scope token \F over individual tokens (ST), and the combined full negation \F (FN).

\subsection{Results}
\label{sec:results}

We present the results for the various benchmarking tasks below.

\begin{table}[t]
    \centering
\begin{tabular}{@{}lllr@{}}
\toprule
  & \multicolumn{2}{c}{POS} &   \\
\cmidrule{2-3}
 Model    & BM & NN & Time \\
\midrule
Stanza \cite{qi-etal-2020-stanza} & 98.3 & 97.9 & --\\
NorELMo$_{30}$ & 98.1 & 97.4 & 8 \\
NorELMo$_{100}$ & 98.0 & 97.4 & 8 \\
mBERT & 98.0 & 97.9  & 245 \\
NB-BERT & \textbf{98.7} & \textbf{98.3} & 244 \\
NorBERT & 98.5 & 98.0 & 238 \\
\bottomrule
\end{tabular}
    \caption{Evaluation scores of the NorLM models on the POS tagging of Bokmål (BM) and Nynorsk (NN) test sets in comparison with other large pre-trained models for Norwegian. Running times in minutes are given for Bokmål.}
    \label{tab:pos_binary}
\end{table}

\begin{table}
\centering
    \begin{tabular}{@{}lccr@{}}
    \toprule
    Model &     Bokmål & Nynorsk & Time\\
     \midrule
    NorELMo$_{30}$ & 79.9 & 75.6 & 2  \\
    NorELMo$_{100}$ & 81.3 & 75.1 & 2 \\
    mBERT & 78.8 & 81.7 & 14  \\
    NB-BERT & \textbf{90.2}  & \textbf{88.6}  &  11 \\
    NorBERT & 85.5 & 82.8 & 9\\
    \bottomrule
    \end{tabular}
    \caption{NER evaluation scores (micro \F) of the NorLM models on the NorNE test set in comparison with other large pre-trained models for Norwegian. Running time is given in minutes for the Bokmål part (on 1 NVIDIA P100 GPU).}
    \label{tab:ner}
\end{table}

\begin{table*}
\newcommand{\impr}[1]{\underline{\textbf{#1}}}
    \centering
    \resizebox{.95\textwidth}{!}{
    \begin{tabular}{@{}lccccccccc@{}}
    \toprule
   & \multicolumn{3}{c}{Spans}  & \multicolumn{1}{c}{Targeted} & \multicolumn{2}{c}{Parsing Graph} & \multicolumn{2}{c}{Sent. Graph} & \\
    \cmidrule(lr){2-4}\cmidrule(lr){5-5}\cmidrule(lr){6-7}\cmidrule(lr){8-9}
    %\cmidrule(lr){10-10}
    Model    & Holder \F & Target \F & Exp. \F & \F & U\F & L\F & NS\F & S\F & Time \\
  \midrule
      Extraction [1] & 42.4 & 31.3 & 31.3 & -- & -- & -- & -- & -- & -- \\
      NorELMo$_{30}$ & 55.1 & 55.3  & 57.2 & \textbf{37.9} & 49.0 & 41.2 & 40.9 & 34.5 & 446\\
      NorELMo$_{100}$ & 58.8 & 55.8 & 56.8 & 37.1 & 49.7 & 41.2  & 41.5 & 34.2 & 434 \\
      mBERT & 57.1  & 55.2  & 56.3 & 34.8   & 48.7  & 38.3 & 40.5  & 31.7 & 444 \\
      NB-BERT & 61.3 & 56.1 & 57.9 & 36.0   & 49.7 & 41.9 & 40.7 & \textbf{34.8} & 404 \\
      NorBERT & \textbf{63.0} & \textbf{56.4} & \textbf{58.1} & 36.9    & \textbf{50.5} & \textbf{42.2} & \textbf{41.0} & \textbf{34.8} & 438 \\
    \bottomrule
    \end{tabular}
    }
    \caption{Average score of NorLM models on fine-grained sentiment (5 runs with set random seeds). \textbf{Bold} denotes the best result on each metric. [1] Span extraction baseline from \newcite{ovrelid-etal-2020-fine}, which uses a BiLSTM CRF with pretrained fastText embeddings.}
    \label{tab:finegrained_results}
\end{table*}

\begin{table}[t]
    \centering
\begin{tabular}{@{}ll@{}}
\toprule
Model & \F \\
\midrule
NorELMo$_{30}$ & 75.0 \\
NorELMo$_{100}$ & 75.0  \\
mBERT  & 67.7  \\
NB-BERT & \textbf{83.9} \\
NorBERT &  77.1  \\
\bottomrule
\end{tabular}
    \caption{\F scores for the different LMs models on the binary sentiment classification test set.}
    \label{tab:sent_binary}
\end{table}

\begin{table}
\centering
    \resizebox{\columnwidth}{!}{
\begin{tabular}{@{}lrrrr@{}}
\toprule
Model & CUE        & ST   & FN  & Time \\
\midrule
     NorELMo$_{30}$        & 91.7   & 80.6 & 63.8 & 428\\
     NorELMo$_{100}$       & 92.2  & 81.3 & 65.5 & 407\\
     mBERT                 & \textbf{92.8} & \textbf{84.0} & \textbf{65.9} & 353 \\
     NB-BERT               & 92.4   & 83.1 & 63.5 & 342\\
     NorBERT               & 92.1  & 83.6 & 65.5 & 426\\
\bottomrule
\end{tabular}
}
    \caption{Results of our negation parser, augmenting the features with
    token representations from each language model. The results are averaged over 5 runs.
    }
    \label{tab:neg_results}
\end{table}

\paragraph{POS tagging}

As can be seen from Table~\ref{tab:pos_binary},  NorBERT outperforms mBERT on both tasks: on POS tagging for Bokmål by 5 percentage points and 1 percentage point for Nynorsk. NorBERT is almost on par with NB-BERT on POS tagging.
NorELMo models are outperformed by NB-BERT and NorBERT, but are on par with mBERT in POS tagging. Note that their adaptation to the tasks (extracting token embeddings and learning a classifier) takes 30x less time than with the BERT models. 

\begin{figure*}[t]
    \centering
    \includegraphics[width=0.45\linewidth]{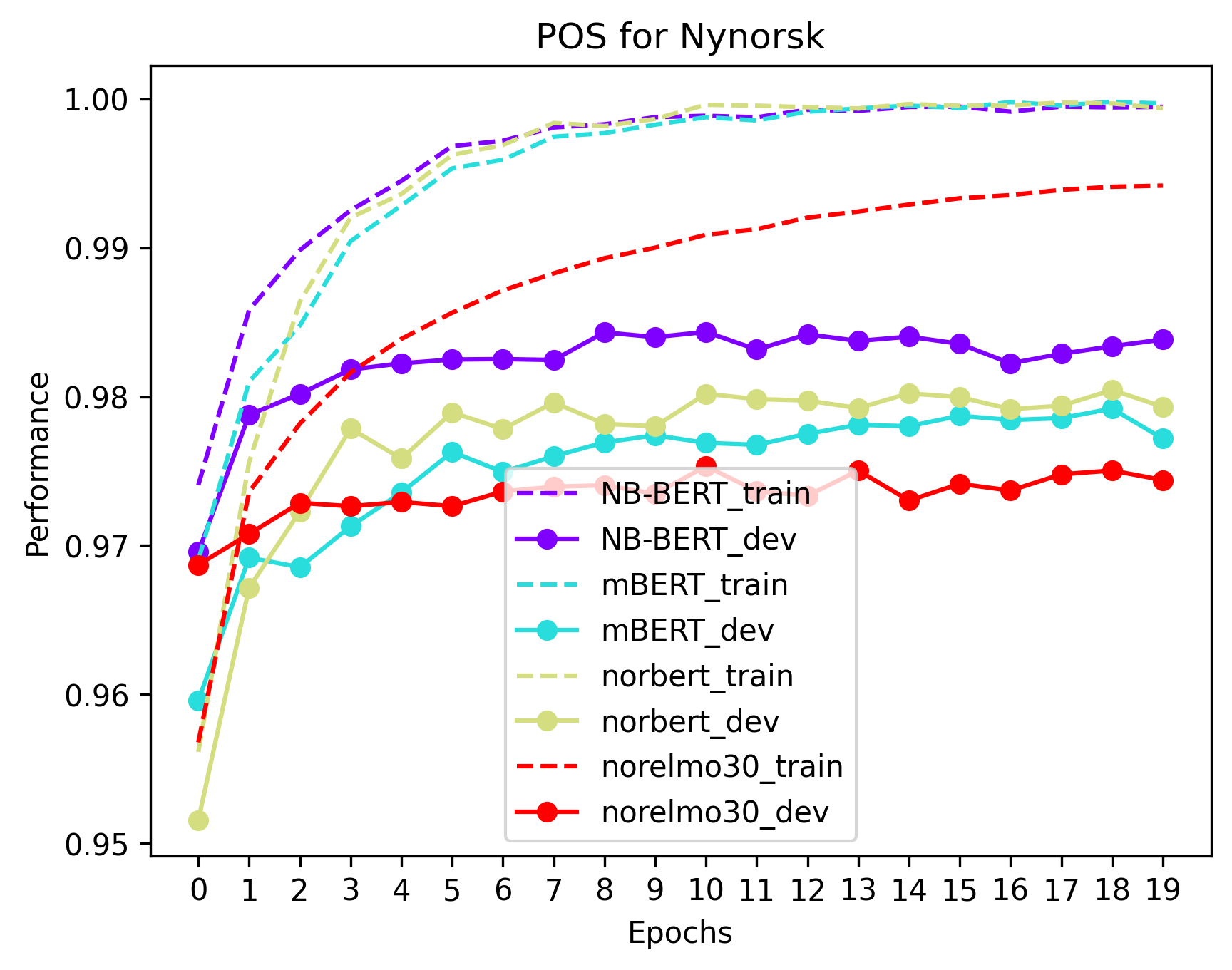}
    \includegraphics[width=0.45\linewidth]{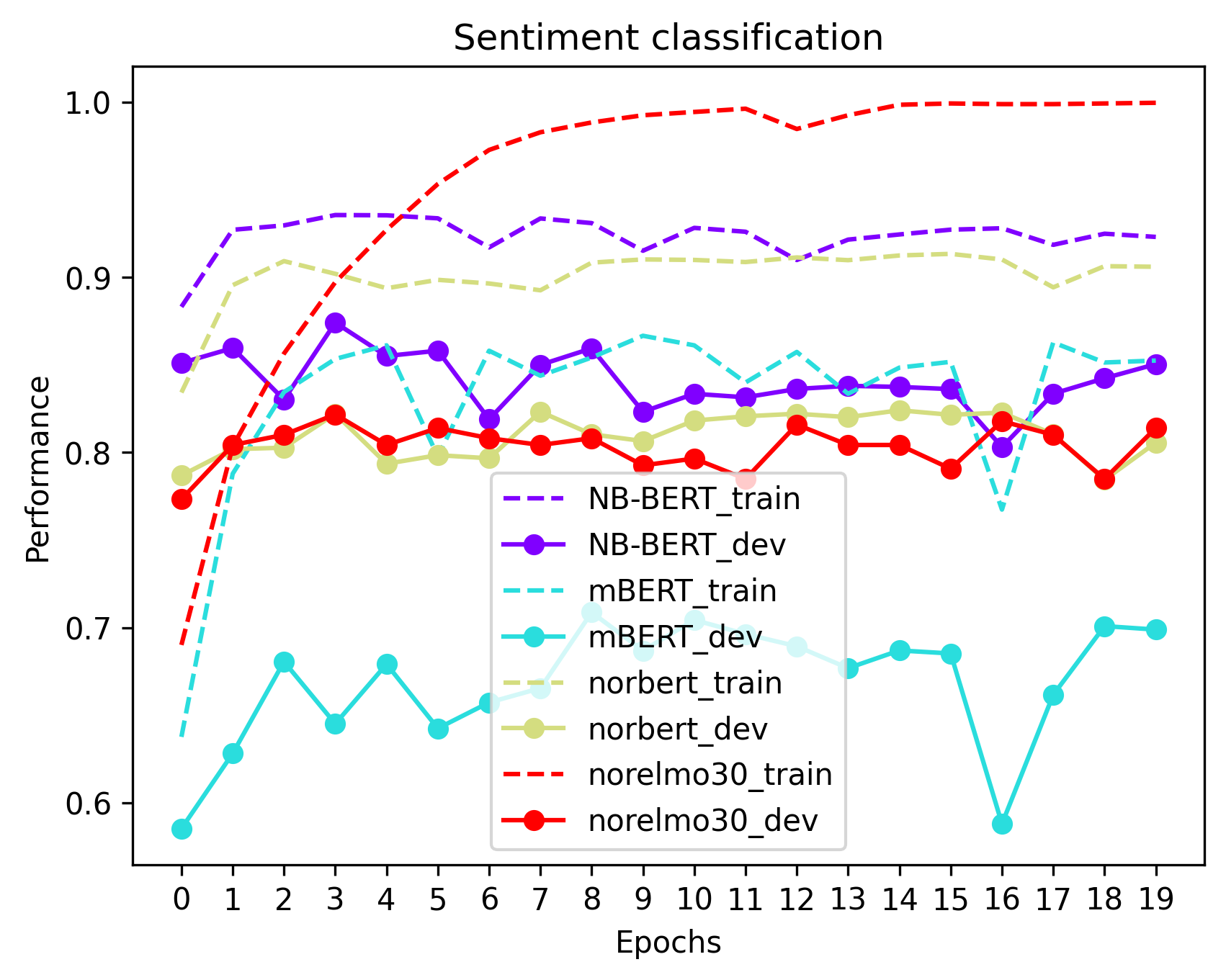}
    \caption{Per-epoch performance on training and development data for two of the tasks. 
    Left: accuracy for POS tagging (Norwegian Nynorsk). Right: \F for binary sentiment classification.}
    \label{fig:epochs}
\end{figure*}

See Figure~\ref{fig:epochs} for the examples of training dynamics of the Nynorsk model.

\paragraph{Named entity recognition}
Table~\ref{tab:ner} shows the performance on the NER task. NB-BERT is the best on both Bokmål and Nynorsk, closely followed by NorBERT. Unsurprisingly, mBERT falls behind all the models trained for Norwegian, when evaluated on Bokmål data. With Nynorsk, it manages to outperform NorELMo. Bokmål is presumably dominant in the training corpora of both. However, in the course of fine-tuning, mBERT seems to be able to adapt to the specifics of Nynorsk. Since our ELMo setup did not include the fine-tuning step, the NorELMo models' adaptation abilities were limited by what can be learned from contextualised token embeddings produced by a frozen model. Still, when used on the data more similar to the training corpus (Bokmål), ELMo achieves competitive results even without any fine-tuning. 

In terms of computational efficiency, the adaptation of ELMo models to this task requires 6x less time than mBERT or NB-BERT and 4x less time than NorBERT. Note also that the NorBERT model takes less time to fine-tune than the NB-BERT model (although the number of epochs was exactly the same), because of a smaller vocabulary, and thus less parameters in the model. Again, in this case an NLP practitioner has a rich spectrum of tools to choose from, depending on whether speed or performance on the downstream task is prioritised.

\paragraph{Fine-grained sentiment analysis}

Table \ref{tab:finegrained_results} shows that NorBERT outperforms mBERT on all metrics and NB-BERT on all but S\F, although the differences between NorBERT and NB-BERT are generally small. 

On this task the NorELMo models generally outperform mBERT as well. However, unlike in the previous tasks, the running times here are similar for BERT and ELMo models, since no fine-tuning was applied (the same is true for negation detection).  We furthermore compare with the previous best model  \cite{ovrelid-etal-2020-fine}, a span extraction model which uses a single-layer Bidirectional LSTM with Conditional Random Field inference, and an embedding layer initialized with fastText vectors trained on the NoWaC corpus. All approaches using language models outperform the previous baseline by a large margin on the span extraction tasks.\footnote{\newcite{ovrelid-etal-2020-fine} only perform span extraction. Therefore, it is not possible to compare the other metrics.} NorBERT, in particular, achieves improvements of 20.6 percentage points on Holder \F (24.9 and 25.8 on Target and Exp. \F, respectively). 

\paragraph{Binary sentiment classification}

Table~\ref{tab:sent_binary} shows that NorBERT outperforms mBERT by 9.4 percentage points on sentiment analysis. However, it seems that in binary sentiment classification the sheer amount of training data starts to show its benefits, and NB-BERT outperforms NorBERT by 6.8 points. NorELMo models outperform mBERT by 7.3 points.

Figure~\ref{fig:epochs} shows the training dynamics of the models.

\paragraph{Negation detection}
 From Table \ref{tab:neg_results} we can see that mBERT gives the best overall results, followed by NorBERT and NorELMo$_{100}$. NB-BERT and NorELMo$_{30}$ perform worse than the others on Scope token \F (ST) and full negation \F (FN), while all models perform similarly at cue-level \F (CUE). We hypothesise that the structural similarity of negation across many of the pretraining languages gives mBERT an advantage, but it is still surprising that it outperforms NB-BERT and NorBERT.

\section{Future plans}
In the future, separate Bokmål and Nynorsk BERT models are planned, and we further expect to train and evaluate models with a higher number of epochs over the training corpus. While we plan to develop additional monolingual Norwegian models based on other contextualised LM architectures beyond BERT and ELMo, we would also be interested to explore the usefulness of multilingual models restricted to Scandinavian languages.  Further streamlining of the  benchmarking process, in terms of both data access and computation of metrics, is something we also want to address in future work.

In addition, the ready availability of a highly optimised software stack on multiple HPC systems (published as part of NorLM) may contribute to other researchers developing very large contextualised language models for additional languages or language variants, e.g.\ domain- or application-specific sub-corpora. We hope that more pre-trained NLP models for Norwegian from both academy and industry will be openly released, making it possible to study the interplay between training corpora sizes, hyperparameters, pre-preprocessing decisions and performance in different tasks. At the same time, given the resource demands and sustainability issues related to training such models, we believe it will be important to coordinate efforts and we hope to collaborate closely with other players moving forward. 

\section{Summary}

This paper has described the first outcomes of NorLM, an initiative coordinated by the Language Technology Group at the University of Oslo seeking to provide Norwegian (and Nordic) large-scale contextualised language models, while simultaneously focusing on maintaining a re-usable software environment for model development on national and Nordic HPC infrastructure. We have here described the training and testing of NorELMo and NorBERT -- the first large-scale monolingual LMs for Norwegian. We have benchmarked the models across a wide array of Norwegian NLP tasks, also comparing to the multilingual mBERT model and another large-scale LM for Norwegian developed in parallel work, NB-BERT, trained on large amounts of text from historical sources. The results show that while the monolingual models tend to yield better results, which particular model ranks first varies across tasks. This underscores the importance of building an ecosystem of diversified models, accompanied by systematic benchmarking.

\section*{Acknowledgements}
The NorLM resources are being developed on the Norwegian national super-computing services operated by UNINETT Sigma2, the National Infrastructure for High Performance Computing and Data Storage in Norway, as well as on the Finnish national supercomputing facilities operated by the CSC IT Center for Science. Software provisioning was financially supported through the European EOSC-Nordic project; data preparation and evaluation were supported by the SANT project (Sentiment Analysis for Norwegian Text), funded by the Research Council of Norway (grant number 270908). 
We are indebted to all funding agencies involved, the University of Oslo, and the Norwegian tax payer.

\bibliographystyle{acl_natbib}
\bibliography{anthology,norbert}
\end{document}